\documentclass{article} 
\usepackage{iclr2021_conference,times}

\usepackage{amsmath,amsfonts,bm,dsfont,amsthm}

\def\eqref#1{equation~\ref{#1}}

\def\1{\bm{1}}

\DeclareMathAlphabet{\mathsfit}{\encodingdefault}{\sfdefault}{m}{sl}
\SetMathAlphabet{\mathsfit}{bold}{\encodingdefault}{\sfdefault}{bx}{n}

\newcommand{\E}{\mathbb{E}}

\newcommand{\indic}{\mathds{1}}

\DeclareMathOperator*{\argmax}{arg\,max}

\newcommand*{\mydprime}{^{\prime\prime}\mkern-1.2mu}

\newtheorem{theorem}{Theorem}[section]

\newtheorem{proposition}[theorem]{Proposition}

\usepackage{xcolor}
\usepackage{listings}

\usepackage{xparse}

\NewDocumentCommand{\codeword}{v}{%
\texttt{\textcolor{black}{#1}}%
}

\usepackage{amssymb}

\usepackage{hyperref}
\usepackage{cleveref}
\usepackage{url}
\usepackage{graphicx}
\usepackage{booktabs}
\usepackage{longtable}
\usepackage{subcaption}
\usepackage{float}
\usepackage[ruled,vlined]{algorithm2e}
\usepackage[section]{placeins}

\title{Planning from Pixels using Inverse Dynamics Models}

\author{Keiran Paster\\
Department of Computer Science\\
University of Toronto, Vector Institute\\
\texttt{keirp@cs.toronto.edu} \\
\And
Sheila A. McIlraith \& Jimmy Ba \\
Department of Computer Science \\
University of Toronto, Vector Institute\\
\texttt{\{sheila, jba\}@cs.toronto.edu} \\
}

\newcommand{\gone}[1]{\textcolor{green}{}}

\newcommand{\hideit}[1]{}

\newcommand{\algoshort}{GLAMOR}
\newcommand{\algoname}{Goal-conditioned Latent Action MOdels for RL}

\iclrfinalcopy 
\begin{document}

\maketitle

\begin{abstract}
Learning task-agnostic dynamics models in high-dimensional observation spaces can be challenging for model-based RL agents. We propose a novel way to learn latent world models by learning to predict sequences of future actions conditioned on task completion. These task-conditioned models adaptively focus modeling capacity on task-relevant dynamics, while simultaneously serving as an effective heuristic for planning with sparse rewards. We evaluate our method on challenging visual goal completion tasks and show a substantial increase in performance compared to prior model-free approaches.
\end{abstract}

\section{Introduction}

Deep reinforcement learning has proven to be a powerful and effective framework for solving a diversity of challenging decision-making problems \citep{alphazero, dota}. However these algorithms are typically trained to maximize a single reward function, ignoring information that is not directly relevant to the associated task at hand. This way of learning is in stark contrast to how humans learn \citep{humans}. Without being prompted by a specific task, humans can still explore their environment, practice achieving imaginary goals, and in so doing learn about the dynamics of the environment. When subsequently presented with a novel task, humans can utilize this learned knowledge to bootstrap learning --- a property we would like our artificial agents to have. In this work, we investigate one way to bridge this gap by learning world models \citep{worldmodels} that enable the realization of previously unseen tasks.

By modeling the task-agnostic dynamics of an environment, an agent can make predictions about how its own actions may affect the environment state without the need for additional samples from the environment. Prior work has shown that by using powerful function approximators to model environment dynamics, training an agent entirely within its own world models can result in large gains in sample efficiency \citep{worldmodels}. However, 
learning world models that are both accurate and general has largely remained elusive, 
with these models experiencing many performance issues in the multi-task setting. 

The main reason for poor performance is the so-called \textit{planning horizon dilemma} \citep{benchmarking}: accurately modeling dynamics over a long horizon is necessary to accurately estimate rewards, but performance is often poor when planning over long sequences due to the accumulation of errors. These modeling errors are especially prevalent in high-dimensional observation spaces where loss functions that operate on pixels may focus model capacity on task-irrelevant features \citep{simple}. Recent work \citep{dreamer, muzero} has attempted to side-step these issues by learning a world model in a latent space and propagating gradients over multiple time-steps. While these methods are able to learn accurate world models and achieve high performance on benchmark tasks, their representations are usually trained with task-specific information such as rewards, encouraging the model to focus on tracking task-relevant features but compromising
their ability to generalize to new tasks.

In this work, we propose to learn powerful, latent world models that can predict environment dynamics when planning for a distribution of tasks. The main contributions of our paper are three-fold:
we propose to learn a latent world model conditioned on a goal; 
we train our latent representation to model inverse dynamics --- sequences of actions that take the agent from one state to another, rather than training it to capture information about reward; and
we show that by combining our inverse dynamics model and a prior over action sequences, we can quickly construct plans that maximize the probability of reaching a goal state. We evaluate our world model on a diverse distribution of challenging visual goals in Atari games and the Deepmind Control Suite \citep{controlsuite} to assess both its accuracy and sample efficiency. We find that when planning in our latent world model, our agent outperforms prior, model-free methods across most tasks, while providing an order of magnitude better sample efficiency on some tasks.

\begin{figure}
	\centering
	\includegraphics[width=380pt,height=172pt]{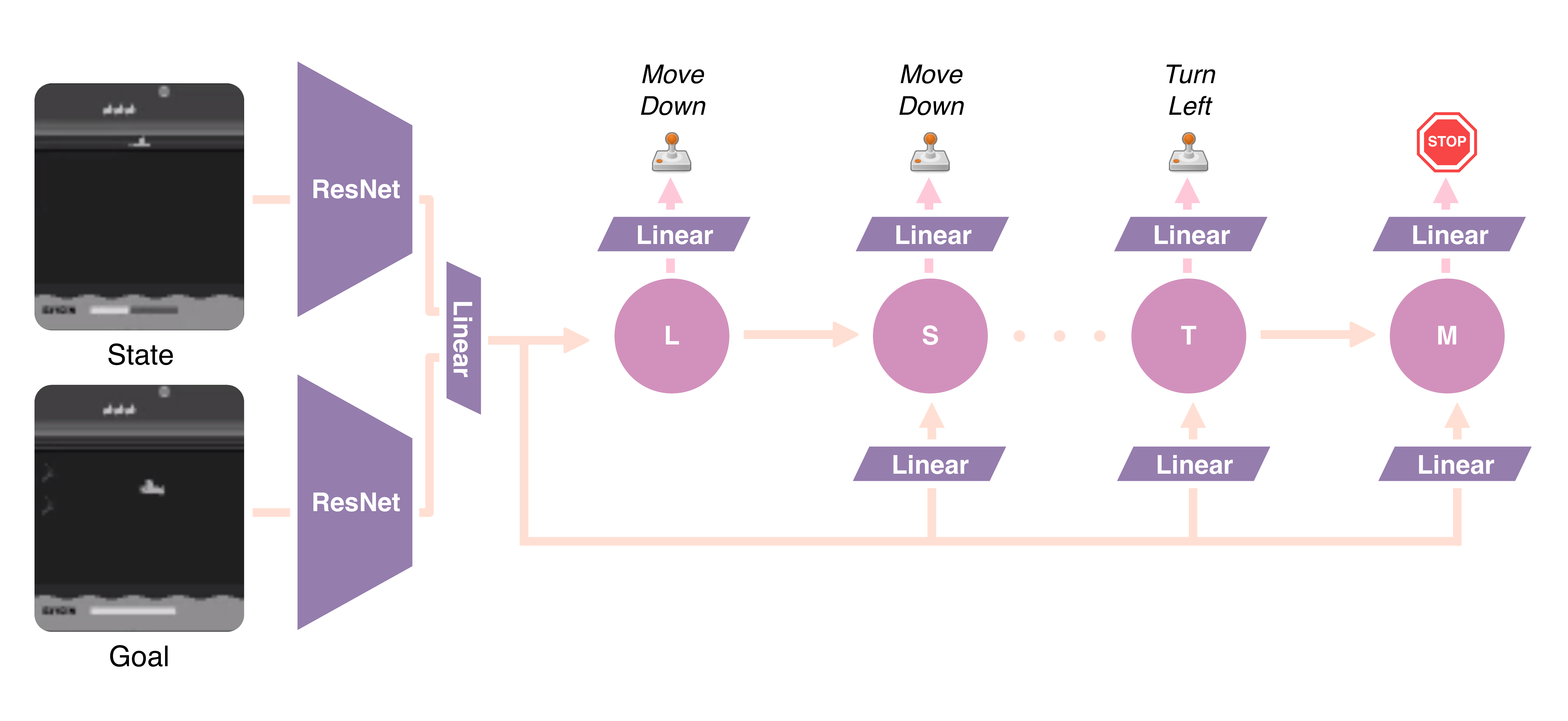}
	\vspace{-0.1in}
	\caption{The network architecture for the inverse dynamics model used in \algoshort{}. ResNets are used to encode state features and an LSTM predicts the action sequence.}
	\label{fig:idmodel}
\end{figure}

\section{Related Work}

Model-based RL has typically focused on learning powerful forward dynamics models, which are trained to predict the next state given the current state and action. In works such as \citep{simple}, these models are trained to predict the next state in observation space - often by minimizing L2 distance. While the performance of these algorithms in the low data regime is often strong, they can struggle to reach the asymptotic performance of model-free methods \citep{dreamer}. An alternative approach is to learn a forward model in a latent space, which may be able to avoid modeling irrelevant features and better optimize for long-term consistency. These latent spaces can be trained to maximize mutual information with the observations \citep{dreamer, planet} or even task-specific quantities like the reward, value, or policy \citep{muzero}. Using a learned forward model, there are several ways that an agent could create a policy. 



While forward dynamics models map a state and action to the next state, an inverse dynamics model maps two subsequent states to an action. Inverse dynamics models have been used in various ways in sequential decision making. In exploration, inverse dynamics serves as a way to learn representations of the controllable aspects of the state \citep{idexplore}. In imitation learning, inverse dynamics models can be used to map a sequence of states to the actions needed to imitate the trajectory \citep{stone}. \citet{transfer} use inverse dynamics models to translate actions taken in a simulated environment to the real world. 

Recently, there has been an emergence of work \citep[e.g.,][]{gcsl, udrlintro, udrlexp} highlighting the relationship between imitation learning and reinforcement learning. Specifically, rather than learn to map states and actions to reward, as is typical in reinforcement learning, \citet{udrlexp} train a model to predict actions given a state and an outcome, which could be the amount of reward the agent is to collect within a certain amount of time. \citet{gcsl} use a similar idea, predicting actions conditioned on an initial state, a goal state, and the amount of time left to achieve the goal. As explored in Appendix \ref{section:gcsl}, these methods are perhaps the nearest neighbors to our algorithm.


In our paper, we tackle a visual goal-completion task due to its generality and ability to generate tasks with no domain knowledge. Reinforcement learning with multiple goals has been studied since \citet{kaelbling}. Most agents that are trained to achieve multiple goals are trained with off-policy reinforcement learning combined with a form of hindsight relabeling \citep{her}, where trajectories that do not achieve the desired goal are relabeled as a successful trajectory that achieves the goal that was actually reached. \citet{her} uses value-based reinforcement learning with a reward based on the euclidean distance between physical objects, which is only possible with access to an object-oriented representation of the state. In environments with high-dimensional observation spaces, goal-achievement rewards are more difficult to design. \citet{rig} use a VAE \citep{vae} trained on observations to construct a latent space and uses distances in the latent space for a reward. These distances, however, may contain features that are uncontrollable or irrelevant. \citet{discern} attempt to solve this issue by framing the goal-achievement task as maximizing the mutual information between the goal and achieved state $I(s_g,s_T)$. Our method differs from these approaches since we aim simply to maximize an indicator reward $\indic(s_T = s_g)$ and do not explicitly learn a value or Q-function.
\section{Method}

\subsection{Problem Formulation}

Reinforcement learning is a framework in which an agent acts in an unknown environment and adapts based on its experience. We model the problem with an MDP, defined as the tuple $(S, A, T, R, \gamma)$. $S$ is a set of states; $A$ is a set of actions; the transition probabilities $T: S \times A \times S \to [0, 1]$ defines the probability of the environment transitioning from state $s$ to $s'$ given that the agent acts with action $a$; the reward function $R: S \times A \times S \to \mathbb{R}$ maps a state-action transition to a real number; and $0 \leq \gamma \leq 1$ is the discount factor, which controls how much an agent should prefer rewards sooner rather than later. An agent acts in the MDP with a policy $\pi: S\times A \to [0, 1]$, which determines the probability of the agent taking action $a$ while in state $s$.

The expected return of a policy is denoted:
\begin{equation}
    J_\text{RL}(\pi) = \E_{\tau \sim P(\tau|\pi)}\left[\sum_t \gamma^t R(s_t, a_t, s'_t)\right],
\end{equation}
that is the averaged discounted future rewards for trajectories $\tau = \{(s_t, a_t)\}_{t=1}^T$ of states and actions sampled from the policy. A reinforcement learning agent's objective is to find the optimal policy $\pi^* = \argmax_\pi J_\text{RL}(\pi)$ that maximizes the expected return.




In goal-conditioned reinforcement learning, an agent's objective is to find a policy that maximizes this return over the distribution of goals $g \sim p(g)$ when acting with a policy that is now also conditioned on $g$. In our work, $g \in S$ and we consider goal achievement rewards of the form $R_g(s) = \indic(s = g)$. Additionally, we consider a trajectory to be complete when any $R_g(s_t) = 1$ and denote this time-step $t = T$. With these rewards, an optimal goal-achieving agent maximizes:

\begin{equation}
    J(\pi) = \E_{g \sim p(g)}[\E_{s_T \sim p(s_T|\pi_g)}[\gamma^T R_g(s_T)]].
\end{equation}
Note that unlike prior works, we consider both the probability of goal achievement as well as the length of the trajectory $T$ in our objective.

\subsection{Planning}

We consider the problem of finding an optimal action sequence $a_1, \ldots, a_{k-1}$ to maximize expected return $J(s, g, a_1, \ldots, a_{k-1})$:
%
%
\begin{equation}
    J(s, g, a_1, \ldots, a_{k-1}) = \E_{s_k \sim p(s_k|s, a_1, \ldots, a_{k-1})}[\gamma^k r_g(s_k)] = \gamma^T p(s_k=g|s, a_1, \ldots, a_{k-1})
\end{equation}
Thus, the optimal action sequence is found by solving the following optimization problem:
\begin{equation}
    \label{equation:max_prob}
    a_1^*, \ldots, a_{k-1}^* = \argmax_{a_1, \ldots, a_{k-1}} \gamma^k p(s_k = g | s_1, a_1, \ldots, a_{k-1})
\end{equation}
Even with access to a perfect model of $p(s_k = g | s_1, a_1, \ldots, a_{k-1})$, solving this optimization may be difficult. 
In many environments, the number of action sequences that reach the goal are vastly outnumbered by the action sequences that do not. Without a heuristic or reward-shaping, there is little hope of solving this problem in a reasonable amount of time.


\subsection{\algoshort{}: \algoname}

Inspired by sequence modeling in NLP, we propose to rewrite \autoref{equation:max_prob} in a way that permits factoring across the actions in the action sequence. By factoring, planning in our model can use the heuristic search algorithms that enable sampling high quality language sequences that are hundreds of tokens long.
%
%
%
%
First, note that\footnote{Here we assume that all goals have a non-zero probability of being achieved under the training distribution.}\footnote{The notation $a_{<i}$ denotes $a_1, \ldots, a_{i-1}$.}:
\begin{equation}
    p(s_k = g | s_1, a_1, \ldots, a_{k-1}) \propto \prod_{i = 1}^{k-1} \frac{p(s_k = g | s_1, a_{< i}, a_i)}{p(s_k = g | s_1, a_{< i})}
\end{equation}

Let $z(s_1, g, a_{<i}, a_i) \triangleq \frac{p(s_k = g | s_1, a_{< i}, a_i)}{p(s_k = g | s_1, a_{< i})}$. Intuitively, these terms are equal to the relative gain in probability of reaching state $g$ conditioned on taking action $a_i$ versus the marginal probability of reaching the goal without conditioning on that action. These terms provide useful information\footnote{By upper-bounding each term $z$, any individual factor being small bounds the rest of the score and allows for early pruning of bad action sequences. An upper-bound can be achieved by assuming some support over all actions in the training policy, which is a common requirement for the convergence of many RL algorithms.} that can guide search towards high scoring action sequences when constructing a plan.



To learn the values of the $z(s_1, g, a_{<i}, a_i)$, we use Bayes' rule to show that we can equivalently learn two auto-regressive behavioral models:
%
%
%
%
%
\begin{align}
    \frac{p(s_k = g | s_1, a_{\leq i})}{p(s_k = g | s_1, a_{\leq i - 1})} &= \frac{p(a_i|s_k = g, s_1, a_{\leq i - 1})p(s_k = g, s_1, a_{\leq i - 1})p(s_1, a_{\leq i - 1})}{p(s_1, a_{\leq i})p(s_k = g, s_1, a_{\leq i - 1})}\\
    &= \frac{p(a_i|s_1, s_k = g, a_{\leq i - 1})}{p(a_i|s_1, a_{\leq i - 1})}
\end{align}
%
%
%
We refer to $p(a_1, \ldots, a_k|s_1, s_k = g)$ as the \textit{inverse dynamics model} and $p(a_1, \ldots, a_k|s_1)$ as the \textit{action prior}. Using these models, we can find an optimal plan by optimizing the following objective:
\begin{equation}
    \label{equation:open_loop}
    a_1^*, \ldots, a_{k-1}^* = \argmax_{a_1, \ldots, a_{k-1}} \gamma^k \frac{p(a_1, \ldots, a_{k-1}|s_1, s_k = g)}{p(a_1, \ldots, a_{k-1}|s_1)}
\end{equation}
\subsection{Learning Inverse Dynamics Model and Action Prior}

We parameterize both the inverse dynamics model and the action prior model with LSTMs \citep{lstm} parameterized by $\theta$ and $\phi$. Please refer to \autoref{section:exp_implementation} for more information about our architecture.

From the training data we construct a new dataset $\mathcal{D}$ of $(s_1, a_1, \ldots, a_{k-1}, s_k)$ tuples for all possible combinations of $s_1$ and $s_k$. Note that this is similar to hindsight relabeling \citep{her}. In training, we find parameters to maximize the likelihood of the training data using AdamW \citep{adamw} to optimize the following loss, where $\alpha$ controls the relative weight of the losses when the two models share parameters:
\begin{equation}
    \begin{aligned}
    \label{equation:loss}
    L(\theta, \phi) = &\E_{(s, a_1, \ldots, a_{k-1}, s_k) \sim \mathcal{D}}[-\log p_\theta(a_1, \ldots, a_{k-1}|s_1, s_k)]\\ &+ \alpha \cdot \E_{(s, a_1, \ldots, a_{k-1}) \sim \mathcal{D}}[-\log p_\phi(a_1, \ldots, a_{k-1}|s_1)].
    \end{aligned}
\end{equation}
While there are many heuristic search algorithms that could be used to optimize \autoref{equation:open_loop}, we opt to use random shooting (RS) \citep{rs} due to ease of parallelization and strong empirical performance. RS samples $N$ candidate action sequences, evaluates the value of the sequences using the learned models, and takes the first action of the winning action sequence. Rather than sampling action sequences uniformly, we sample them auto-regressively using a Boltzmann distribution of the scores $z$.

In order to be able to re-plan at every step, our model must be able to model action sequences with variable length. By predicting variable length action sequences using end-tokens, our model is able to plan to both maximize goal-achievement probability and find a shortest path.

\begin{algorithm}[H]
\SetAlgoLined
 Initialize inverse dynamics model $p_\theta(a_1, \ldots, a_{k-1}|s_1, s_k)$\;
 Initialize action prior model $p_\phi(a_1, \ldots, a_{k-1}|s_1)$\;
 Initialize dataset $\mathcal{D}((s, a))$\;
 \For{\text{training iteration} $k = 1, 2, 3,$ \ldots}{
    Sample $t$ trajectories with training policy $\pi_k$ and goals sampled from $p(g)$\;
    Add tuples $(s_1, a_1, \ldots, a_{k-1}, s_k)$ to dataset $\mathcal{D}$\;
    Update $\theta, \phi$ by descending the gradient of \autoref{equation:loss}\;
 }
 \caption{\algoshort{}}
\end{algorithm}

\subsection{Data Collection}

While training data need not be collected on-policy, we note that using an arbitrary data-collection policy may result in a causally-incorrect model \citep{cpm}. In our experiments, we use our open-loop planner to generate a sequence of actions given just the initial state $s_1$ and a sampled goal $g$ and follow this sequence for the rest of the episode. For a discussion of the types of errors that may occur with a causally incorrect model, see Appendix \ref{section:causal}.

For exploration and in order to ensure support over all actions, we use an $\epsilon$-greedy exploration policy where the $i$th action in the plan is taken with probability $1-\epsilon$ and an action is randomly sampled with probability $\epsilon$. We decay $\epsilon$ to a small final value over time. We store trajectories in a circular replay buffer, meaning dataset $\mathcal{D}$ contains trajectories from multiple past versions of the planner.

\subsection{Comparison to Prior Methods}

Prior methods for goal-conditioned RL include DISCERN \citep{discern} and Goal-Conditioned Supervised Learning (GCSL) \citep{gcsl}. DISCERN learns a goal discriminator and Q-network and rewards the agent for ending up in states where its discriminator can easily guess the desired goal. GCSL learns a policy with supervised learning by iteratively imitating its own relabeled trajectories. While \algoshort{} outputs a sequence of actions using an LSTM, these prior works only output a single action for the current time-step.

While \algoshort{} shows strong empirical performance in benchmark tasks, we believe that there are several other key advantages over prior goal-conditioned RL algorithms:

\begin{itemize}
    \item \algoshort{} learns a world model, which is more flexible than the Q-networks or policies that prior methods use. For example, despite only being trained to find shortest paths to a goal state, the latent world model learned in \algoshort{} contains information about many different paths to the goal. One implication is that during planning, the agent can be modified to select action sequences that reach the goal at a specific time-step. This is experimentally verified in \autoref{section:planning}.
    \item Q-learning is well known \citep{offline} to perform poorly with purely off-policy data and GCSL relies on many iterations of training and data-collection to converge to an optimal policy. As shown in Appendix \ref{section:train_policy}, \algoshort{} performs well even with a completely random training policy, often still achieving SOTA performance. Strong off-policy performance may enable the use of offline datasets or more sophisticated exploration policies.
    \item GCSL does not estimate the prior probability of taking an action sequence and therefore fails to converge to an optimal policy when action sequences that are optimal for one goal may land the agent in another goal state. By using the action prior to disentangle the probability of reaching a goal from the probability of taking an action sequence, \algoshort{} avoids this issue. For a more detailed comparison of \algoshort{} and GCSL, see section Appendix \ref{section:gcsl}.
\end{itemize}
\newpage
\begin{figure}
  \vspace{-0.2in}
  \label{fig:algo_average}
    \centering
    \includegraphics[width=0.99\linewidth]{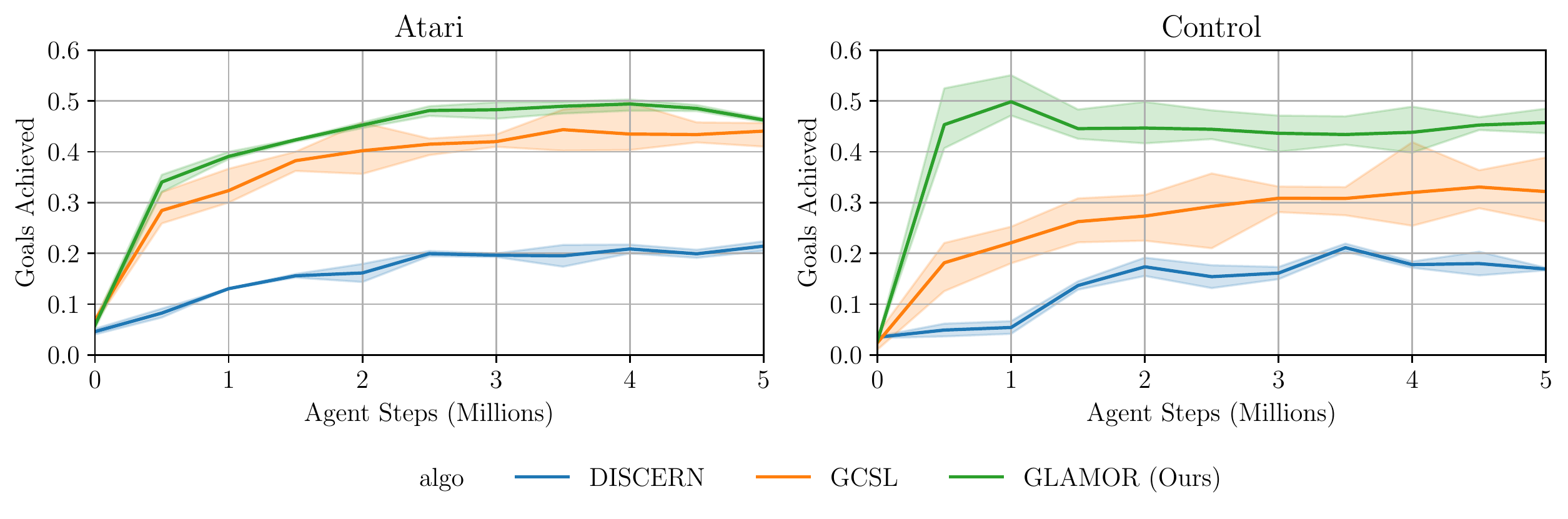}
    \vspace{-0.1in}
\caption{Both in Atari and on tasks from the Deepmind Control Suite, \algoshort{} outperforms prior methods. The goal achievement rate is averaged over all games / control tasks and over three seeds. See \autoref{fig:atari_curves} and \autoref{fig:control_curves} in the appendix for more detailed training curves.
\label{fig:tables}}
\vspace{-0.2in}
\end{figure}
\section{Experiments}

In our experiments, we aim to answer the following questions:

\begin{itemize}
    \item How accurate is our world model? Can the agent plan using its model to achieve a diverse set of goals even in high-dimensional domains?

    \item How sample efficient are our agents? In the low sample regime, is it better to use \algoshort{} or use model-free reinforcement learning? 
    
    \item How effective is our planning procedure? Can the planner discover how to achieve goals with shortest paths or paths with a specific length?

\end{itemize}

\begin{table}[b]
    \scriptsize
\vspace{-0.2in}
\caption{Goal Achievement Rates at 500k Agent Steps in Atari
\vspace{0.in}}
\begin{subtable}[t]{0.5\textwidth}
\begin{tabular}[t]{llll}
\toprule
Atari &         GLAMOR (Ours) &      DISCERN &                  GCSL \\
\midrule
Bowling          &           0.14 (0.07) &  0.11 (0.09) &  \textbf{0.21 (0.11)} \\
Boxing           &  \textbf{0.06 (0.04)} &  0.01 (0.01) &           0.01 (0.02) \\
Breakout         &  \textbf{0.12 (0.05)} &  0.02 (0.02) &           0.04 (0.03) \\
Frostbite        &  \textbf{0.35 (0.06)} &  0.04 (0.04) &           0.20 (0.05) \\
Montezuma &           0.13 (0.07) &  0.03 (0.02) &  \textbf{0.30 (0.07)} \\
MsPacman         &           0.28 (0.07) &  0.04 (0.02) &  \textbf{0.30 (0.05)} \\
Pitfall          &  \textbf{0.40 (0.08)} &  0.07 (0.04) &           0.26 (0.07) \\
Pong             &  \textbf{0.44 (0.20)} &  0.21 (0.08) &           0.28 (0.06) \\
PrivateEye       &  \textbf{0.26 (0.07)} &  0.04 (0.04) &           0.16 (0.06) \\
Qbert            &  \textbf{0.83 (0.03)} &  0.26 (0.04) &           0.56 (0.17) \\
Riverraid        &  \textbf{0.80 (0.05)} &  0.27 (0.03) &           0.79 (0.09) \\
Seaquest         &  \textbf{0.57 (0.06)} &  0.04 (0.04) &           0.48 (0.08) \\
Skiing           &           0.30 (0.09) &  0.02 (0.02) &  \textbf{0.38 (0.08)} \\
Tennis           &  \textbf{0.10 (0.04)} &  0.00 (0.00) &           0.05 (0.03) \\
\bottomrule
\end{tabular}
\caption{}
\end{subtable}
\hfill
\begin{subtable}[t]{0.5\textwidth}
\begin{tabular}[t]{llll}
\toprule
DM Control &         GLAMOR (Ours) &               DISCERN &                  GCSL \\
\midrule
ball\_in\_cup      &  \textbf{0.20 (0.08)} &           0.01 (0.02) &           0.03 (0.04) \\
cartpole       &  \textbf{0.97 (0.03)} &           0.01 (0.01) &           0.09 (0.02) \\
finger            &  \textbf{0.09 (0.05)} &           0.01 (0.01) &           0.03 (0.02) \\
manipulator &  \textbf{0.00 (0.00)} &  \textbf{0.00 (0.00)} &  \textbf{0.00 (0.00)} \\
pendulum       &  \textbf{0.83 (0.09)} &           0.20 (0.05) &           0.14 (0.05) \\
point\_mass        &           0.76 (0.16) &           0.03 (0.02) &  \textbf{0.82 (0.05)} \\
reacher           &  \textbf{0.27 (0.08)} &           0.08 (0.06) &           0.07 (0.04) \\

\bottomrule
\end{tabular}
\caption{}
\end{subtable}
\label{table:500kcontrol}
\vspace{-0.2in}
\label{table:500k}
\end{table}

\subsection{Environments}

We evaluate our method on two types of environments: Atari games and control tasks in the Deepmind Control Suite \citep{controlsuite}. In both environments, visual observations are converted to grayscale and down-scaled to $(80, 104)$ pixels. In order to incorporate historical information, we opt to concatenate the four most recent frames to form an observation, as introduced by \citet{dqn}. However, we opt to use only a single frame for specifying visual goals. We calculate goal achievement in both environments using extracted low dimensional features.



\noindent{\sc Arcade Learning Environment (ALE).} We run experiments on a subset of the available ALE games, chosen by the availability of labeling methods and the suitability of the games to the goal-achieving tasks. In all games we use a frame-skip of $4$, and all Atari environments used a random number of initial noops and sticky actions to introduce stochasticity.


\noindent{\sc Deepmind Control Suite.} We choose to use the subset of the Deepmind Control Suite \citep{controlsuite} chosen by \citet{discern}. We also adopt the same method of discretization, discretizing most of the $A$-dimensional continuous control tasks into $3^A$ actions. For \codeword{manipulator}, we adopt DISCERN's diagonal discretization. In \codeword{point mass} we apply a frame-skip of $4$ frames, with no frame-skip for any other control environment.

\subsection{Implementation Details}
\label{section:exp_implementation}

We parameterize our models with two convolutional encoders, similar to the large model in \citet{impala}, and an LSTM model \citep{lstm} for action prediction. See \autoref{fig:idmodel} for a visual description of our architecture. We opt to share parameters only in the encoders for the inverse dynamics model and action prior. In DISCERN and GCSL, we represent the time remaining with a periodic representation $(\sin(2\pi t/T), \cos(2\pi t/T))$. Detailed training hyperparameters are available at Appendix \ref{section:hyperparams}.

The code for training agents on both Atari and DM Control Suite along with evaluation code will be released upon publication.

\subsection{Baselines}

We evaluate our method against both GCSL and DISCERN. We chose DISCERN due to its reported high performance on our benchmark tasks, and GCSL due to its similarity to our own algorithm. GCSL was implemented within our code-base, ensuring that any differences are due to algorithmic differences rather than implementation details. Our implementation was checked against publicly available code. For DISCERN, due to the lack of available source code, we made a best effort attempt to reproduce the algorithm. We built our implementation on top of rlpyt's \citep{rlpyt} Rainbow implementation \citep{rainbow}. All algorithms use the same encoder architecture, and hyperparameters were fixed between GCSL and \algoshort{}.

 

\begin{figure}[t]
\vspace{-0.2in}
\centering
\begin{subfigure}[b]{.5\textwidth}
  \centering
  \includegraphics[width=0.8\linewidth]{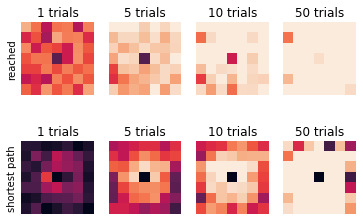}
  \vspace{0.42in}
  \caption{Goal achievement rates for a 7x7 grid-world.}
  \label{fig:compute}
\end{subfigure}%
\begin{subfigure}[b]{.5\textwidth}
  \centering
  \includegraphics[width=0.9\linewidth]{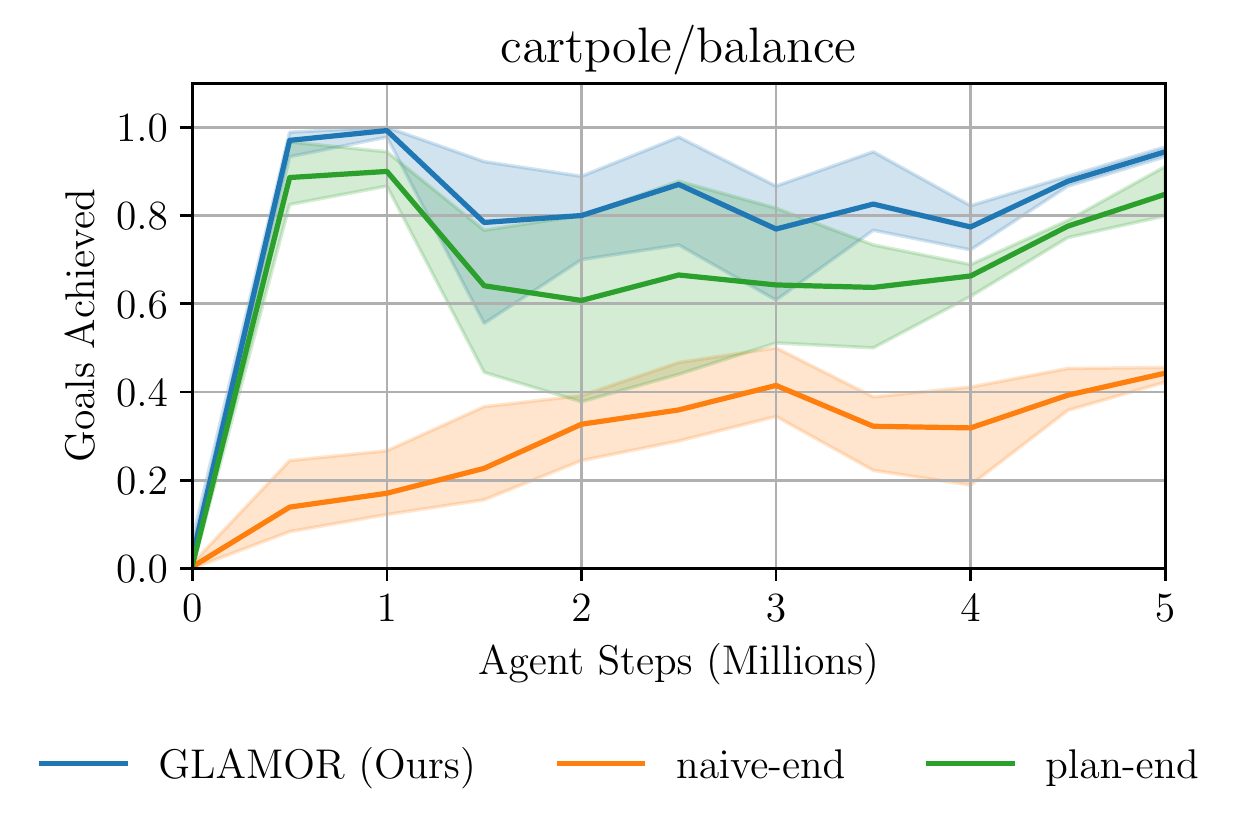}
  \caption{Goal achievement rates for different termination strategies.}
  \label{fig:termination}
\end{subfigure}
\caption{(a) The agent starts in the center and must travel to the goal tile. Top shows the rate at which the agent eventually achieved the goal and bottom shows the rate at which the agent achieved the goal with the shortest available path. The amount of compute used for planning is shown on the x-axis. As the planning budget increases, both the number of successfully reached goals and the number of goals achieved optimally improves substantially. Brighter means a higher achievement rate. (b) In ``naive-end'', the agent greedily tries to take a shortest path to the goal for $T$ timesteps and is evaluated at the end. In ``plan-end'', the agent explicitely constructs a plan to achieve the goal state at the end of its trajectory. \algoshort{} (Ours) can choose to terminate its episode early. \vspace{-0.1in}}
\label{fig:test}
\end{figure}

\subsection{Evaluation}

To evaluate the degree to which an agent achieved its desired goal, we extract the positions of the various entities in the scene. For Atari, we extract state entity labels using code from \citet{labels}. In the Deepmind Control Suite environments, we use the position information from the non-visual representation. Information from both of these agents is kept private from the agents at both train and evaluation time. As in DISCERN, a goal is considered to be achieved if the positions of the entities are within 10\% of their feasible range. Positions are evaluated at the end of the episode ($T = 50$ in Atari and $T = 100$ in control tasks) or when the agent decides to terminate the episode. In order to inform \algoshort{} of the time remaining until evaluation, we limit the length of the plan in the planner after $t$ steps to $T - t$.

Agents are evaluated on a set of $30$ fixed goals per environment. These goals are generated similarly to the \textit{diverse} goal buffer described in \citet{discern}: goals are repeatedly sampled and a goal is only added to the buffer if it is farther away from the closest goal in feature space than the goal that it is replacing is. We found this procedure to generate a good coverage of possible goals for evaluation.




\subsection{Achieving Visually-Specified Goals}

In order to test our model's accuracy, we evaluate our agent's performance in achieving visually-specified goals in all of our test environments. \autoref{fig:tables} shows that by planning in its latent world model, our agent learns to achieve goals with at least as much accuracy as DISCERN and GCSL in 17 out of 21 tasks, often achieving as much as twice as many goals. See \autoref{fig:atari_curves} and \autoref{fig:control_curves} in the appendix for individual training curves for each environment. Despite not being explicitly trained to only focus on controllable features as in \citet{discern}, \autoref{fig:control_frames} shows our agent learning to control the position of its finger even when it can't exactly match the orientation of the spinner. Overall, we conclude that since planning under our agent's world model results in a high goal-achievement rate relative to previous state-of-the-art algorithms, our model is sufficiently accurate. Videos of \algoshort{} agents achieving goals in all environments are available at \url{https://sites.google.com/view/glamor-paper}.

\subsection{Sample Efficiency}

Model-based reinforcement learning is known for having better sample efficiency than model-free algorithms. We found this to be true for \algoshort{} as well. In \autoref{table:500k}, we show the performance of our algorithm trained with only 500k agent steps. \algoshort{} achieves decidedly more goals than DISCERN and GCSL. Of particular note is that DISCERN is sample inefficient\footnote{\citet{discern} trained DISCERN for 200M steps in their paper.}, learning to control almost none of the control tasks at 500k steps while \algoshort{} has already converged.

\subsection{Planning}
\label{section:planning}

We ran several experiments to test properties of our planning procedure.

\subsubsection{Planning Compute}

One benefit of model-based reinforcement learning is that the agent can improve its performance or even adapt its policy at test-time simply by changing its planning procedure. In order to test the performance of \algoshort{} with various levels of compute used to construct a plan, we used an empty grid-world environment where shortest-paths are known. We evaluate the percentage of trajectories that reach the goal in an optimal amount of time for each goal. \autoref{fig:compute} shows a heatmap of the optimal goal-completion rates for different amounts of compute used during planning. As the amount of samples in the planner is increased, both the rate of achieving goals and the rate of achieving goals with an optimal path increase.

\subsubsection{Early Termination}

We also test the flexibility of the learned model by slightly changing the task. While \algoshort{} is trained to find shortest paths to the goal state, we ran an additional experiment evaluating its performance if the agent is evaluated on its ability to get to the goal state in exactly $T$ steps. For this experiment, we use the \codeword{cartpole/balance} environment, since terminating early when the cart is in the correct position is significanly easier than manipulating it to reach that position at a specific time. In \autoref{fig:termination}, we show that while early termination achieves the highest goal-achievement rate, explicitly planning to achieve the goal at the end of the trajectory outperforms naively taking shortest paths towards the goal state for $T$ time-steps. All three experiments used the exact same learned model and the only difference is in the planning procedure, where ``plan-end'' does not sample the termination token until the last time-step. Videos of the different termination strategies are also available on the website.

\begin{figure}[t]
\centering
\includegraphics[width=1\linewidth]{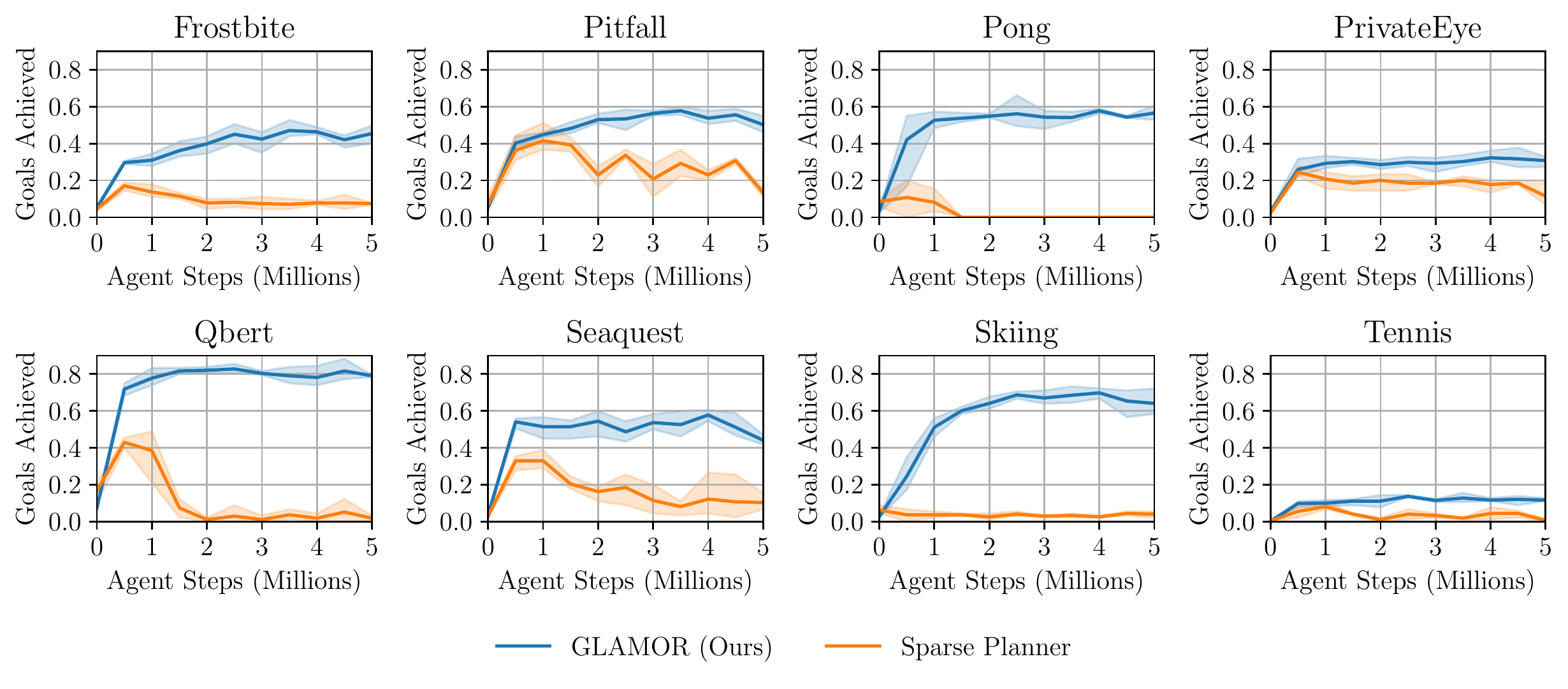}
\caption{Using intermediate information to guide the planning process helps \algoshort{} achieve more goals than when it only looks at the estimated probability of reaching a goal at the end of the episode.}
\label{fig:sparse_planner}
\end{figure}

\subsubsection{Sparse Planning}

Prior works such as MuZero \citep{muzero} and the Predictron \citep{predictron} also learn latent dynamics models and are similar to \algoshort{} in that they predict rewards given a starting state and action sequence. The reward (probability of reaching the goal state) predicted in \algoshort{} however is in the form of the inverse dynamics model and action prior, as shown in \autoref{equation:open_loop}, and the intermediate probabilities predicted by the auto-regressive models are used to guide the search for a good action sequence. To test the contribution of this guidance, we compare \algoshort{} to a version where candidate action sequences are generated randomly and the highest scoring one under \autoref{equation:open_loop} is selected. \autoref{fig:sparse_planner} shows that performance is significantly worse in the tested environments.

\section{Conclusion}

We have presented a novel way to learn latent world models by modeling inverse dynamics. These models learn to track task-relevant dynamics for a diverse distribution of tasks and provide a strong heuristic that enables efficient planning. We also demonstrate strong performance in both the low and high sample regime on 21 challenging visual benchmark tasks.

Goal-achievement tasks already have significant practical value. A next step is to extend \algoshort{} to general reward functions. \algoshort{} also learns a non-reactive policy. While combining non-reactive planning with Model Predictive Control has proven to be sufficient in many benchmark tasks, a natural future direction is to account for these types of stochastic environments.

\section*{Acknowledgements}

We gratefully acknowledge funding from the Natural Sciences and Engineering Research Council of Canada (NSERC), the Canada CIFAR AI Chairs Program, and Microsoft Research. Resources used in preparing this research were provided, in part, by the Province of Ontario, the Government of Canada through CIFAR, and companies sponsoring the Vector Institute for Artificial Intelligence (\url{www.vectorinstitute.ai/partners}).

\bibliography{iclr2021_conference}

\begin{thebibliography}{33}
\providecommand{\natexlab}[1]{#1}
\providecommand{\url}[1]{\texttt{#1}}
\expandafter\ifx\csname urlstyle\endcsname\relax
  \providecommand{\doi}[1]{doi: #1}\else
  \providecommand{\doi}{doi: \begingroup \urlstyle{rm}\Url}\fi

\bibitem[Anand et~al.(2019)Anand, Racah, Ozair, Bengio, C{\^{o}}t{\'{e}}, and
  Hjelm]{labels}
Ankesh Anand, Evan Racah, Sherjil Ozair, Yoshua Bengio, Marc{-}Alexandre
  C{\^{o}}t{\'{e}}, and R.~Devon Hjelm.
\newblock Unsupervised state representation learning in atari.
\newblock In \emph{Advances in Neural Information Processing Systems 32: Annual
  Conference on Neural Information Processing Systems 2019, NeurIPS 2019, 8-14
  December 2019, Vancouver, BC, Canada}, pp.\  8766--8779, 2019.

\bibitem[Andrychowicz et~al.(2017)Andrychowicz, Crow, Ray, Schneider, Fong,
  Welinder, McGrew, Tobin, Abbeel, and Zaremba]{her}
Marcin Andrychowicz, Dwight Crow, Alex Ray, Jonas Schneider, Rachel Fong, Peter
  Welinder, Bob McGrew, Josh Tobin, Pieter Abbeel, and Wojciech Zaremba.
\newblock Hindsight experience replay.
\newblock In \emph{Advances in Neural Information Processing Systems 30: Annual
  Conference on Neural Information Processing Systems 2017, 4-9 December 2017,
  Long Beach, CA, {USA}}, pp.\  5048--5058, 2017.

\bibitem[Berner et~al.(2019)Berner, Brockman, Chan, Cheung, Debiak, Dennison,
  Farhi, Fischer, Hashme, Hesse, J{\'{o}}zefowicz, Gray, Olsson, Pachocki,
  Petrov, de~Oliveira~Pinto, Raiman, Salimans, Schlatter, Schneider, Sidor,
  Sutskever, Tang, Wolski, and Zhang]{dota}
Christopher Berner, Greg Brockman, Brooke Chan, Vicki Cheung, Przemyslaw
  Debiak, Christy Dennison, David Farhi, Quirin Fischer, Shariq Hashme,
  Christopher Hesse, Rafal J{\'{o}}zefowicz, Scott Gray, Catherine Olsson,
  Jakub Pachocki, Michael Petrov, Henrique~Pond{\'{e}} de~Oliveira~Pinto,
  Jonathan Raiman, Tim Salimans, Jeremy Schlatter, Jonas Schneider, Szymon
  Sidor, Ilya Sutskever, Jie Tang, Filip Wolski, and Susan Zhang.
\newblock Dota 2 with large scale deep reinforcement learning.
\newblock \emph{arXiv preprint arXiv: 1912.06680}, 2019.

\bibitem[Christiano et~al.(2016)Christiano, Shah, Mordatch, Schneider,
  Blackwell, Tobin, Abbeel, and Zaremba]{transfer}
Paul~F. Christiano, Zain Shah, Igor Mordatch, Jonas Schneider, Trevor
  Blackwell, Joshua Tobin, Pieter Abbeel, and Wojciech Zaremba.
\newblock Transfer from simulation to real world through learning deep inverse
  dynamics model.
\newblock \emph{arXiv preprint arXiv: 1610.03518}, 2016.

\bibitem[Espeholt et~al.(2018)Espeholt, Soyer, Munos, Simonyan, Mnih, Ward,
  Doron, Firoiu, Harley, Dunning, Legg, and Kavukcuoglu]{impala}
Lasse Espeholt, Hubert Soyer, R{\'{e}}mi Munos, Karen Simonyan, Volodymyr Mnih,
  Tom Ward, Yotam Doron, Vlad Firoiu, Tim Harley, Iain Dunning, Shane Legg, and
  Koray Kavukcuoglu.
\newblock {IMPALA:} scalable distributed deep-rl with importance weighted
  actor-learner architectures.
\newblock In \emph{Proceedings of the 35th International Conference on Machine
  Learning, {ICML} 2018, Stockholmsm{\"{a}}ssan, Stockholm, Sweden, July 10-15,
  2018}, volume~80 of \emph{Proceedings of Machine Learning Research}, pp.\
  1406--1415. {PMLR}, 2018.

\bibitem[Fujimoto et~al.(2019)Fujimoto, Meger, and Precup]{offline}
Scott Fujimoto, David Meger, and Doina Precup.
\newblock Off-policy deep reinforcement learning without exploration.
\newblock In Kamalika Chaudhuri and Ruslan Salakhutdinov (eds.),
  \emph{Proceedings of the 36th International Conference on Machine Learning,
  {ICML} 2019, 9-15 June 2019, Long Beach, California, {USA}}, volume~97 of
  \emph{Proceedings of Machine Learning Research}, pp.\  2052--2062. {PMLR},
  2019.
\newblock URL \url{http://proceedings.mlr.press/v97/fujimoto19a.html}.

\bibitem[Ghosh et~al.(2020)Ghosh, Gupta, Reddy, Fu, Devin, Eysenbach, and
  Levine]{gcsl}
Dibya Ghosh, Abhishek Gupta, Ashwin Reddy, Justin Fu, Coline Devin, Benjamin
  Eysenbach, and Sergey Levine.
\newblock Learning to reach goals via iterated supervised learning.
\newblock \emph{arXiv preprint arXiv: 1912.06088}, 2020.

\bibitem[Ha \& Schmidhuber(2018)Ha and Schmidhuber]{worldmodels}
David Ha and J{\"{u}}rgen Schmidhuber.
\newblock Recurrent world models facilitate policy evolution.
\newblock In Samy Bengio, Hanna~M. Wallach, Hugo Larochelle, Kristen Grauman,
  Nicol{\`{o}} Cesa{-}Bianchi, and Roman Garnett (eds.), \emph{Advances in
  Neural Information Processing Systems 31: Annual Conference on Neural
  Information Processing Systems 2018, NeurIPS 2018, 3-8 December 2018,
  Montr{\'{e}}al, Canada}, pp.\  2455--2467, 2018.

\bibitem[Hafner et~al.(2019)Hafner, Lillicrap, Fischer, Villegas, Ha, Lee, and
  Davidson]{planet}
Danijar Hafner, Timothy~P. Lillicrap, Ian Fischer, Ruben Villegas, David Ha,
  Honglak Lee, and James Davidson.
\newblock Learning latent dynamics for planning from pixels.
\newblock In \emph{Proceedings of the 36th International Conference on Machine
  Learning, {ICML} 2019, 9-15 June 2019, Long Beach, California, {USA}},
  volume~97 of \emph{Proceedings of Machine Learning Research}, pp.\
  2555--2565. {PMLR}, 2019.

\bibitem[Hafner et~al.(2020)Hafner, Lillicrap, Ba, and Norouzi]{dreamer}
Danijar Hafner, Timothy~P. Lillicrap, Jimmy Ba, and Mohammad Norouzi.
\newblock Dream to control: Learning behaviors by latent imagination.
\newblock In \emph{8th International Conference on Learning Representations,
  {ICLR} 2020, Addis Ababa, Ethiopia, April 26-30, 2020}, 2020.

\bibitem[Hessel et~al.(2018)Hessel, Modayil, van Hasselt, Schaul, Ostrovski,
  Dabney, Horgan, Piot, Azar, and Silver]{rainbow}
Matteo Hessel, Joseph Modayil, Hado van Hasselt, Tom Schaul, Georg Ostrovski,
  Will Dabney, Dan Horgan, Bilal Piot, Mohammad~Gheshlaghi Azar, and David
  Silver.
\newblock Rainbow: Combining improvements in deep reinforcement learning.
\newblock In \emph{Proceedings of the Thirty-Second {AAAI} Conference on
  Artificial Intelligence, (AAAI-18), New Orleans, Louisiana, USA, February
  2-7, 2018}, pp.\  3215--3222. {AAAI} Press, 2018.

\bibitem[Hochreiter \& Schmidhuber(1997)Hochreiter and Schmidhuber]{lstm}
Sepp Hochreiter and Jürgen Schmidhuber.
\newblock Long short-term memory.
\newblock \emph{Neural computation}, 9:\penalty0 1735--80, 12 1997.

\bibitem[Holtzman et~al.(2020)Holtzman, Buys, Du, Forbes, and Choi]{nucleus}
Ari Holtzman, Jan Buys, Li~Du, Maxwell Forbes, and Yejin Choi.
\newblock The curious case of neural text degeneration.
\newblock In \emph{8th International Conference on Learning Representations,
  {ICLR} 2020, Addis Ababa, Ethiopia, April 26-30, 2020}. OpenReview.net, 2020.
\newblock URL \url{https://openreview.net/forum?id=rygGQyrFvH}.

\bibitem[Kaelbling(1993)]{kaelbling}
Leslie~Pack Kaelbling.
\newblock Learning to achieve goals.
\newblock In \emph{Proceedings of the 13th International Joint Conference on
  Artificial Intelligence (IJCAI-93)}, pp.\  1094--1098, 1993.

\bibitem[Kaiser et~al.(2020)Kaiser, Babaeizadeh, Milos, Osinski, Campbell,
  Czechowski, Erhan, Finn, Kozakowski, Levine, Mohiuddin, Sepassi, Tucker, and
  Michalewski]{simple}
Lukasz Kaiser, Mohammad Babaeizadeh, Piotr Milos, Blazej Osinski, Roy~H.
  Campbell, Konrad Czechowski, Dumitru Erhan, Chelsea Finn, Piotr Kozakowski,
  Sergey Levine, Afroz Mohiuddin, Ryan Sepassi, George Tucker, and Henryk
  Michalewski.
\newblock Model based reinforcement learning for atari.
\newblock In \emph{8th International Conference on Learning Representations,
  {ICLR} 2020, Addis Ababa, Ethiopia, April 26-30}, 2020.

\bibitem[Kingma \& Welling(2014)Kingma and Welling]{vae}
Diederik~P. Kingma and Max Welling.
\newblock Auto-encoding variational bayes.
\newblock In Yoshua Bengio and Yann LeCun (eds.), \emph{2nd International
  Conference on Learning Representations, {ICLR} 2014, Banff, AB, Canada, April
  14-16, 2014}, 2014.

\bibitem[Loshchilov \& Hutter(2019)Loshchilov and Hutter]{adamw}
Ilya Loshchilov and Frank Hutter.
\newblock Decoupled weight decay regularization.
\newblock \emph{arXiv preprint arXiv: 1711.05101}, 2019.

\bibitem[Mnih et~al.(2015)Mnih, Kavukcuoglu, Silver, Rusu, Veness, Bellemare,
  Graves, Riedmiller, Fidjeland, Ostrovski, Petersen, Beattie, Sadik,
  Antonoglou, King, Kumaran, Wierstra, Legg, and Hassabis]{dqn}
Volodymyr Mnih, Koray Kavukcuoglu, David Silver, Andrei~A. Rusu, Joel Veness,
  Marc~G. Bellemare, Alex Graves, Martin~A. Riedmiller, Andreas Fidjeland,
  Georg Ostrovski, Stig Petersen, Charles Beattie, Amir Sadik, Ioannis
  Antonoglou, Helen King, Dharshan Kumaran, Daan Wierstra, Shane Legg, and
  Demis Hassabis.
\newblock Human-level control through deep reinforcement learning.
\newblock \emph{Nat.}, 518\penalty0 (7540):\penalty0 529--533, 2015.

\bibitem[Nair et~al.(2018)Nair, Pong, Dalal, Bahl, Lin, and Levine]{rig}
Ashvin Nair, Vitchyr Pong, Murtaza Dalal, Shikhar Bahl, Steven Lin, and Sergey
  Levine.
\newblock Visual reinforcement learning with imagined goals.
\newblock In Samy Bengio, Hanna~M. Wallach, Hugo Larochelle, Kristen Grauman,
  Nicol{\`{o}} Cesa{-}Bianchi, and Roman Garnett (eds.), \emph{Advances in
  Neural Information Processing Systems 31: Annual Conference on Neural
  Information Processing Systems 2018, NeurIPS 2018, 3-8 December 2018,
  Montr{\'{e}}al, Canada}, pp.\  9209--9220, 2018.

\bibitem[Pathak et~al.(2017)Pathak, Agrawal, Efros, and Darrell]{idexplore}
Deepak Pathak, Pulkit Agrawal, Alexei~A. Efros, and Trevor Darrell.
\newblock Curiosity-driven exploration by self-supervised prediction.
\newblock In Doina Precup and Yee~Whye Teh (eds.), \emph{Proceedings of the
  34th International Conference on Machine Learning, {ICML} 2017, Sydney, NSW,
  Australia, 6-11 August 2017}, volume~70 of \emph{Proceedings of Machine
  Learning Research}, pp.\  2778--2787, 2017.

\bibitem[Pavse et~al.(2019)Pavse, Torabi, Hanna, Warnell, and Stone]{stone}
Brahma~S. Pavse, Faraz Torabi, Josiah~P. Hanna, Garrett Warnell, and Peter
  Stone.
\newblock {RIDM:} reinforced inverse dynamics modeling for learning from a
  single observed demonstration.
\newblock \emph{arXiv preprint arXiv: 1906.07372}, 2019.

\bibitem[Rezende et~al.(2020)Rezende, Danihelka, Papamakarios, Ke, Jiang,
  Weber, Gregor, Merzic, Viola, Wang, Mitrovic, Besse, Antonoglou, and
  Buesing]{cpm}
Danilo~J. Rezende, Ivo Danihelka, George Papamakarios, Nan~Rosemary Ke, Ray
  Jiang, Theophane Weber, Karol Gregor, Hamza Merzic, Fabio Viola, Jane Wang,
  Jovana Mitrovic, Frederic Besse, Ioannis Antonoglou, and Lars Buesing.
\newblock Causally correct partial models for reinforcement learning.
\newblock \emph{arXiv preprint arXiv: 2002.02836}, 2020.

\bibitem[Richards(2005)]{rs}
A.~Richards.
\newblock \emph{Robust constrained model predictive control}.
\newblock PhD thesis, Massachusetts Institute of Technology, 2005.

\bibitem[Schmidhuber(2019)]{udrlintro}
Juergen Schmidhuber.
\newblock Reinforcement learning upside down: Don't predict rewards -- just map
  them to actions.
\newblock \emph{arXiv preprint arXiv: 1912.02875}, 2019.

\bibitem[Schrittwieser et~al.(2019)Schrittwieser, Antonoglou, Hubert, Simonyan,
  Sifre, Schmitt, Guez, Lockhart, Hassabis, Graepel, Lillicrap, and
  Silver]{muzero}
Julian Schrittwieser, Ioannis Antonoglou, Thomas Hubert, Karen Simonyan,
  Laurent Sifre, Simon Schmitt, Arthur Guez, Edward Lockhart, Demis Hassabis,
  Thore Graepel, Timothy~P. Lillicrap, and David Silver.
\newblock Mastering atari, go, chess and shogi by planning with a learned
  model.
\newblock \emph{arXiv preprint arXiv: 1911.08265}, 2019.

\bibitem[Silver et~al.(2017{\natexlab{a}})Silver, Hubert, Schrittwieser,
  Antonoglou, Lai, Guez, Lanctot, Sifre, Kumaran, Graepel, Lillicrap, Simonyan,
  and Hassabis]{alphazero}
David Silver, Thomas Hubert, Julian Schrittwieser, Ioannis Antonoglou, Matthew
  Lai, Arthur Guez, Marc Lanctot, Laurent Sifre, Dharshan Kumaran, Thore
  Graepel, Timothy~P. Lillicrap, Karen Simonyan, and Demis Hassabis.
\newblock Mastering chess and shogi by self-play with a general reinforcement
  learning algorithm.
\newblock \emph{arXiv preprint arXiv: 1712.01815}, 2017{\natexlab{a}}.

\bibitem[Silver et~al.(2017{\natexlab{b}})Silver, van Hasselt, Hessel, Schaul,
  Guez, Harley, Dulac{-}Arnold, Reichert, Rabinowitz, Barreto, and
  Degris]{predictron}
David Silver, Hado van Hasselt, Matteo Hessel, Tom Schaul, Arthur Guez, Tim
  Harley, Gabriel Dulac{-}Arnold, David~P. Reichert, Neil~C. Rabinowitz,
  Andr{\'{e}} Barreto, and Thomas Degris.
\newblock The predictron: End-to-end learning and planning.
\newblock In Doina Precup and Yee~Whye Teh (eds.), \emph{Proceedings of the
  34th International Conference on Machine Learning, {ICML} 2017, Sydney, NSW,
  Australia, 6-11 August 2017}, volume~70 of \emph{Proceedings of Machine
  Learning Research}, pp.\  3191--3199. {PMLR}, 2017{\natexlab{b}}.
\newblock URL \url{http://proceedings.mlr.press/v70/silver17a.html}.

\bibitem[Srivastava et~al.(2019)Srivastava, Shyam, Mutz, Jaśkowski, and
  Schmidhuber]{udrlexp}
Rupesh~Kumar Srivastava, Pranav Shyam, Filipe Mutz, Wojciech Jaśkowski, and
  Jürgen Schmidhuber.
\newblock Training agents using upside-down reinforcement learning.
\newblock \emph{arXiv preprint arXiv: 1912.02877}, 2019.

\bibitem[Stooke \& Abbeel(2019)Stooke and Abbeel]{rlpyt}
Adam Stooke and Pieter Abbeel.
\newblock rlpyt: A research code base for deep reinforcement learning in
  pytorch.
\newblock \emph{arXiv preprint arXiv: 1909.01500}, 2019.

\bibitem[Tassa et~al.(2018)Tassa, Doron, Muldal, Erez, Li, de~Las~Casas,
  Budden, Abdolmaleki, Merel, Lefrancq, Lillicrap, and
  Riedmiller]{controlsuite}
Yuval Tassa, Yotam Doron, Alistair Muldal, Tom Erez, Yazhe Li, Diego
  de~Las~Casas, David Budden, Abbas Abdolmaleki, Josh Merel, Andrew Lefrancq,
  Timothy~P. Lillicrap, and Martin~A. Riedmiller.
\newblock Deepmind control suite.
\newblock \emph{arXiv preprint arXiv: 1801.00690}, 2018.

\bibitem[Tenenbaum(2018)]{humans}
Josh Tenenbaum.
\newblock Building machines that learn and think like people.
\newblock In Elisabeth Andr{\'{e}}, Sven Koenig, Mehdi Dastani, and Gita
  Sukthankar (eds.), \emph{Proceedings of the 17th International Conference on
  Autonomous Agents and MultiAgent Systems, {AAMAS} 2018, Stockholm, Sweden,
  July 10-15}, pp.\ ~5. International Foundation for Autonomous Agents and
  Multiagent Systems Richland, SC, {USA}, 2018.

\bibitem[Wang et~al.(2019)Wang, Bao, Clavera, Hoang, Wen, Langlois, Zhang,
  Zhang, Abbeel, and Ba]{benchmarking}
Tingwu Wang, Xuchan Bao, Ignasi Clavera, Jerrick Hoang, Yeming Wen, Eric
  Langlois, Shunshi Zhang, Guodong Zhang, Pieter Abbeel, and Jimmy Ba.
\newblock Benchmarking model-based reinforcement learning.
\newblock \emph{arXiv preprint arXiv: 1907.02057}, 2019.

\bibitem[Warde{-}Farley et~al.(2019)Warde{-}Farley, de~Wiele, Kulkarni,
  Ionescu, Hansen, and Mnih]{discern}
David Warde{-}Farley, Tom~Van de~Wiele, Tejas~D. Kulkarni, Catalin Ionescu,
  Steven Hansen, and Volodymyr Mnih.
\newblock Unsupervised control through non-parametric discriminative rewards.
\newblock In \emph{7th International Conference on Learning Representations,
  {ICLR} 2019, New Orleans, LA, USA, May 6-9}, 2019.

\end{thebibliography}
\bibliographystyle{iclr2021_conference}

\appendix
\section{Appendix}




\subsection{Comparison to Goal Conditional Supervised Learning}
\label{section:gcsl}

\citet{gcsl} propose a similar algorithm where a policy is trained by imitation learning with a maximum-likelihood loss:
\begin{equation}
    L(\pi) = \E_\mathcal{D}[-\log \pi_t(a|s, g, h)]
\end{equation}
Here $h$ is the time remaining until evaluation. GCSL iterates between training a policy to clone the behavior of the previous policy conditioned on achieving goal $g$ and gathering new training data.

The key differences between GCSL and \algoshort{} are:
(i) To act, \algoshort{} solves an optimization problem while GCSL samples directly from the policy;
(ii) \algoshort{} learns both an inverse dynamics model and the action prior while GCSL only learns a policy $\pi(s, a, g, h)$; and (iii) Data for GCSL is collected with a reactive policy while \algoshort{} uses an open-loop plan.

\hideit{
\begin{itemize}
    \item To act, \algoshort{} solves an optimization problem while GCSL samples directly from the policy.
    \item \algoshort{} learns both an inverse dynamics model and the action prior while GCSL only learns a policy $\pi(s, a, g, h)$.
    \item Data for GCSL is collected with a reactive policy while \algoshort{} uses an open-loop plan.
\end{itemize}
}

In our experiments, we found that GCSL performed surprisingly well on visual goal-achievement tasks despite the original implementation by \citet{gcsl} only being tested on control tasks with low-dimensional observations. 


\citet{gcsl} provide a theoretical analysis of GCSL, and show that their behavioral cloning objective is a bound on the RL objective, with looseness induced by both how off-policy the training distribution is and the relabeling step. To illustrate the sub-optimal behavior of GCSL caused by relabeling, consider a simplified setting of a one-step MDP and on-policy training data. 

\begin{proposition}
Let $a_*(g)$ be an action that maximizes the probability $p(s=g|\cdot)$. If there exist two goals $g, g'$ such that $p(g) > 0$, $p(g') > 0$, and $p(s=g|a_*(g)) > p(s=g|a_*(g')) > 0$, then one-step GCSL does not converge to an optimal policy.
\end{proposition}

As an example, consider a game with state space $S = \{1, 2, 3, 4, 5, 6\}$ and action space \{\textit{roll fair die}, \textit{roll loaded die}\}. Rolling the fair die will result in the state being uniformly set to the number that the die lands on, while the loaded die will always land on $1$. The optimal goal-achieving policy is evident: the agent should roll the loaded die with probability one only when the goal is to land on $1$. Otherwise the agent should always roll the fair die. In this example, GCSL will fail to converge but \algoshort{} will find the correct policy.


\begin{proof}
\label{section:gcsl_proof}

We consider a one-step MDP with goal distribution $p(g)$. In this MDP, we assume a deterministic initial state and denote the state the agent transitions to after taking action $a$ as $s$.

For simplicity, we assume the policy $\pi_\theta$ is powerful enough to perfectly match its training distribution when trained with maximum-likelihood. Therefore, GCSL consists of the following iterated steps:

\begin{itemize}
    \item Gather trajectories with $\pi_t$. This gives an empirical distribution of goal-conditioned actions $p_{t+1}(a|s=g) = \frac{p(s=g|a)p_t(a)}{p(s=g)}$.
    \item Distill $p_{t+1}(a|s=g)$ into $\pi_{t+1}(a|g)$.
\end{itemize}

The update to the policy $\pi$ at each iteration depends both on the relative likelihood of transitioning to the goal with the environment dynamics and the probability of taking action $a$, $p_t(a)$. Clearly, $p_t(a)$ is a function of the previous policy $\pi_t$ and the goal distribution $p(g)$:

\begin{equation}
    p_t(a) = \sum_{g'} p(g')\pi_t(a|g')
\end{equation}

Assuming the distillation is exact, the policy evolves like:

\begin{equation}
    \pi_{t+1}(a|s=g) = \frac{p(s=g|a)}{p(s=g)} \sum_{g'} p(g')\pi_t(a|g')
\end{equation}

We then analyze the ratio of the probability of any sub-optimal action $a$ to the probability of the optimal action $a^*$: $\frac{\pi_t(a|g)}{\pi_t(a^*(g)|g)}$.

If there is \textit{goal interference}, goals $g, g'$ exist such that $p(g')p(s=g|a_*(g')) > 0$ and $p(s=g|a_*(g)) > p(s=g|a_*(g')) > 0$. 

Assume $\pi_{t-1}$ is optimal. Then,

\begin{align}
    \frac{\pi_t(a_*(g')|g)}{\pi_t(a_*(g)|g)} &= \frac{p(s=g|a_*(g'))\sum_{g\mydprime} p(g\mydprime)\pi_{t-1}(a_*(g')|g\mydprime)}{p(s=g|a_*(g))\sum_{g\mydprime} p(g\mydprime)\pi_{t-1}(a_*(g)|g\mydprime)}\\
    &\geq \frac{p(s=g|a_*(g'))p(g')}{p(s=g|a_*(g))}\\
    &> 0
\end{align}

Therefore, if $\pi_{t-1}$ is optimal, $\pi_t$ will again be sub-optimal and the policy will never converge.
\end{proof}

\subsection{Causally Correct Models}
\label{section:causal}

\citet{cpm} explore the connection between model-based reinforcement learning and causality. A model is \textit{causally correct} if a learned model $q_\theta(x) \approx p(x)$ with respect to a set of interventions. In model-based RL, a model is trained to predict some aspect of an environment. In order to use the learned model to predict the affect of a new policy in the environment, the model must be causally correct with respect to changes in the policy. \citet{cpm} show that some partial models, including MuZero, are not causally correct with respect to action sequence interventions. 

As an example, consider training a model to predict whether a sequence of actions wins in the game of Simon Says. The model is trained to predict $p(\text{win}|s_0, a_1, \ldots, a_k)$ on data produced by some training policy. If this training policy is good at the game, the conditional probability $p(\text{win}|s_0, a_1, \ldots, a_k) = 1$ for all action sequences, even though using the action sequence blindly in the real game would result in a much lower win-rate. This happens because the true data-generating process is actually dependent on intermediate states $s_1, \ldots, s_k$, which the training policy has access to. By conditioning on some action sequence $s_0, a_1, \ldots, a_k$ without modeling the intermediate states, these states become confounding variables. In order to predict the affect of taking a certain action sequence on the environment, what we really want to do is find $p(\text{win}|s_0, do(a_1, \ldots, a_k))$.

To learn causally correct models with \algoshort{}, we opt to simplify the data-generating process by using training policies that are independent of intermediate states. While this may hurt training in some stochastic environments, we find that in our multi-task setting with relabeling, using non-reactive exploration has a negligible effect. 

\subsection{Experimental Details}
\label{section:hyperparams}

\begin{figure}
	\centering
	\begin{tabular}{l|l}
		\hline
		Hyper-parameter & value\\
		\hline
		optimizer & AdamW\\
		weight-decay & 0.01\\
		normalization & GroupNorm\\
		learning-rate & 5e-4\\
		replay-ratio & 4\\
		eps-steps & 3e5\\
		eps-final & 0.1\\
		min-steps-learn & 5e4\\
		buffer size & 1e6\\
		policy trials & 50\\
		state size & 512\\
		clip-p-actions & -3.15\\
		lstm-hidden-dim & 64\\
		lstm-layers & 1\\
		train tasks & 1000\\
		\hline
	\end{tabular}
	\caption{Hyperparameters used to train \algoshort{}.}
	\label{table:hyperparam}
\end{figure}

\autoref{table:hyperparam} shows the hyperparameters that were used to train our method. While our method is substantially more simple than value-based methods like DISCERN, there are still a few important hyperparameters. We found that tuning the replay ratio is important to balance between sample efficiency and over-fitting in our models. We also find that \algoshort{} works best with a large replay buffer and a large model. We also found that avoiding selecting action sequences which have too low a probability, similar to tricks used in beam search in NLP \citep{nucleus}, increases the performance of our planner. To achieve this, we introduce a hyperparameter \codeword{clip-p-actions}, and only expand an action sequence if the immediate log-probability under the inverse dynamics model is over this value.

\subsection{Ablations}

In order to find which parts of \algoshort{} contribute to its performance, we ran an ablation study.

\subsubsection{Training Policy}
\label{section:train_policy}

\begin{figure}[t]
  	\centering
	\includegraphics[width=1\linewidth]{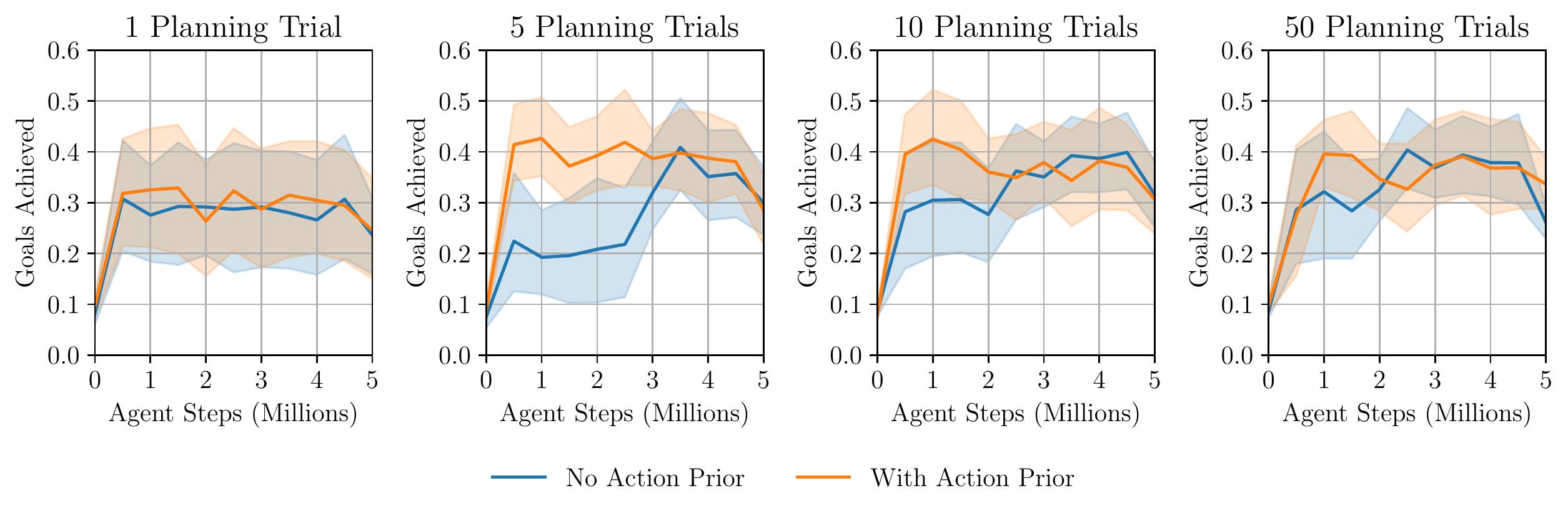}
	\caption{In this experiment, we test how changing the training policy affects performance in Pong. As the amount of compute used in the planner during training increases, so does the performance of the evaluated agent. Using the action prior decreases the variance of the agent's performance.}
	\label{fig:train_ablation}
\end{figure}

In \autoref{fig:train_ablation}, we vary the way we construct the open-loop action sequence that is followed to collect training data. We vary it in two ways: the amount of compute used to create the plan and whether we consider the action prior. While the performance is poor when using only one planning sample, \algoshort{} seems to work well on Pong with as little as 5 samples. More interestingly, disabling the action prior during training seems to increase the variance of \algoshort{} significantly. Without the action prior, the planning procedure will select action sequences that may have shown up more in the training data simply due to the training policy. We hypothesize that this effect can significantly hurt exploration and performance.

\begin{figure}[t]
  	\centering
	\includegraphics[width=1\linewidth]{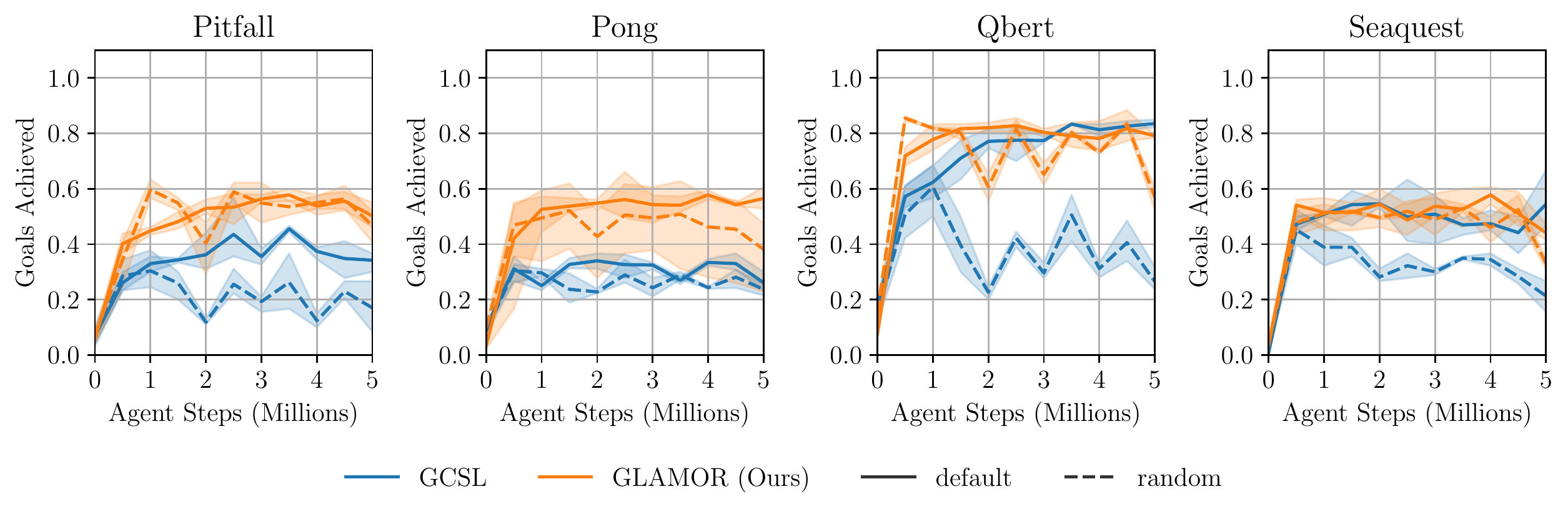}
	\caption{When training agents with off-policy data collected with a random policy, \algoshort{} outperfoms GCSL and can achieve most goals.}
	\label{fig:uniform}
\end{figure}

We also test the performance of both \algoshort{} and GCSL when using a uniform training policy in \autoref{fig:uniform}. We found that \algoshort{} performs very well even when trained on completely off-policy data, while GCSL struggles.


\subsubsection{Planner}

\begin{figure}[t]
  \centering
  \includegraphics[width=1\linewidth]{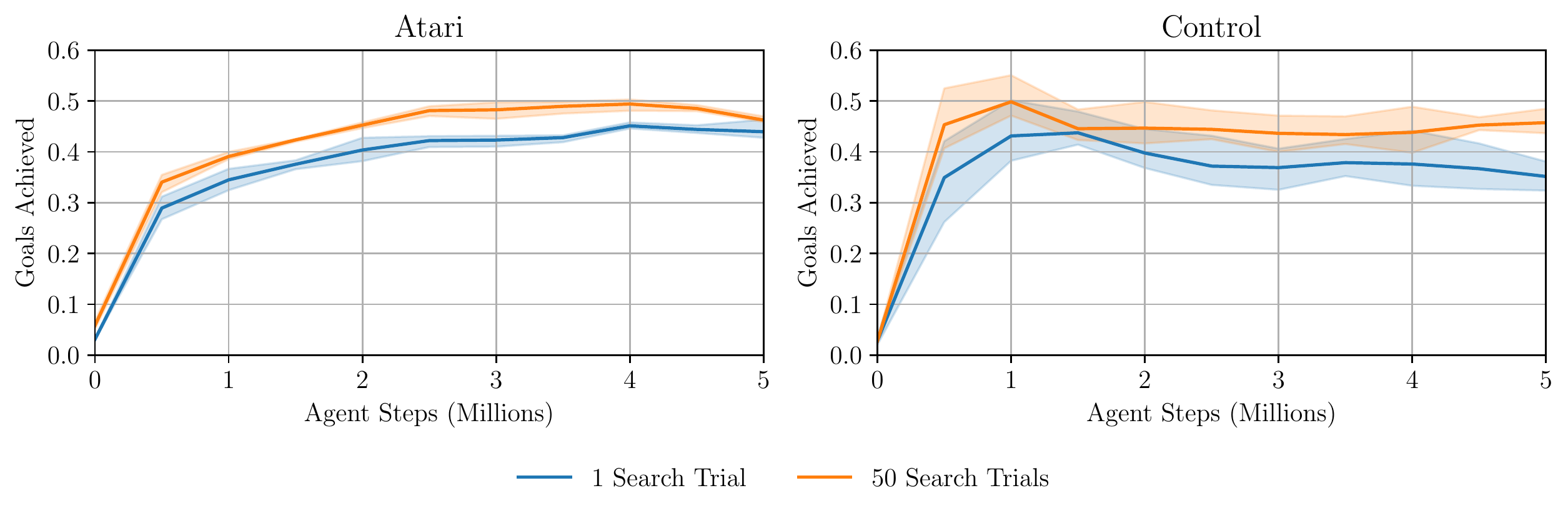}
 
  \caption{Goal achievement rates for DM Control and Atari. Searching for a high scoring action sequence results in more goals achieved. However, even when using compute equivalent to a model-free agent, \algoshort{} performs remarkably well.}
  \label{fig:trials_average}
\end{figure}

We also evaluate whether the planner is necessary at test-time to achieve strong goal-achievement performance. To test this, we ran an experiment where the planner simply takes 1 trajectory sample and takes the first action\footnote{Note that this is exactly equivalent to lowering the planning compute in our implementation.}.  \autoref{fig:trials_average} shows that while the agent still surprisingly achieves many goals in this setting, using the planner results in a stronger policy. We interperate this result as showing that the heuristic search guided by the factored inverse dynamics and action prior is strong and additional compute simply chooses a plan among already good options.

\subsection{Additional Figures}

\begin{figure}[t]
\centering
\includegraphics[width=1\linewidth]{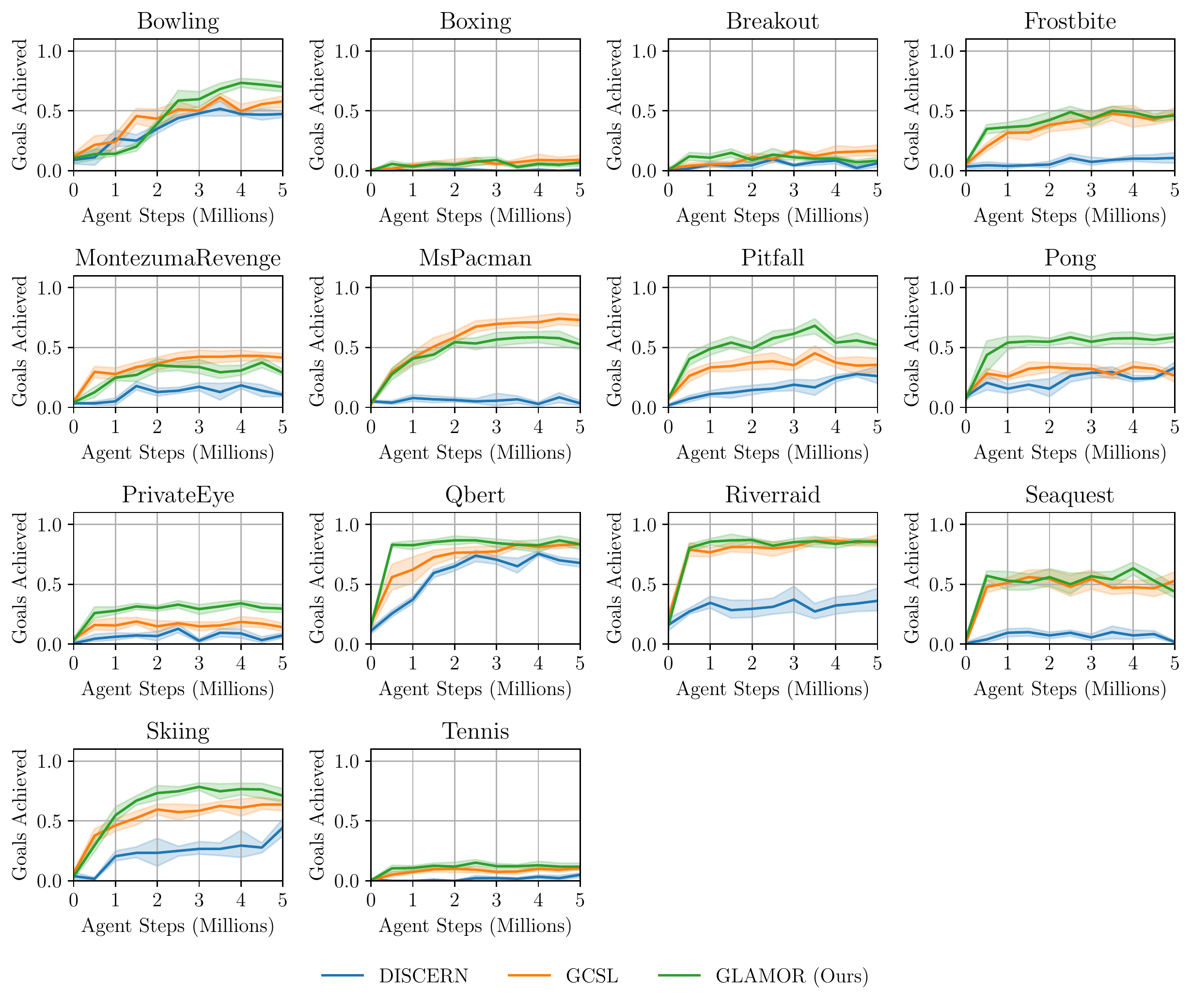}
\caption{Training Curves for Atari Tasks. \algoshort{} achieves more goals (often with many fewer steps) than both GCSL and DISCERN.}
\label{fig:atari_curves}
\end{figure}

\begin{figure}[t]
\centering
\includegraphics[width=1\linewidth]{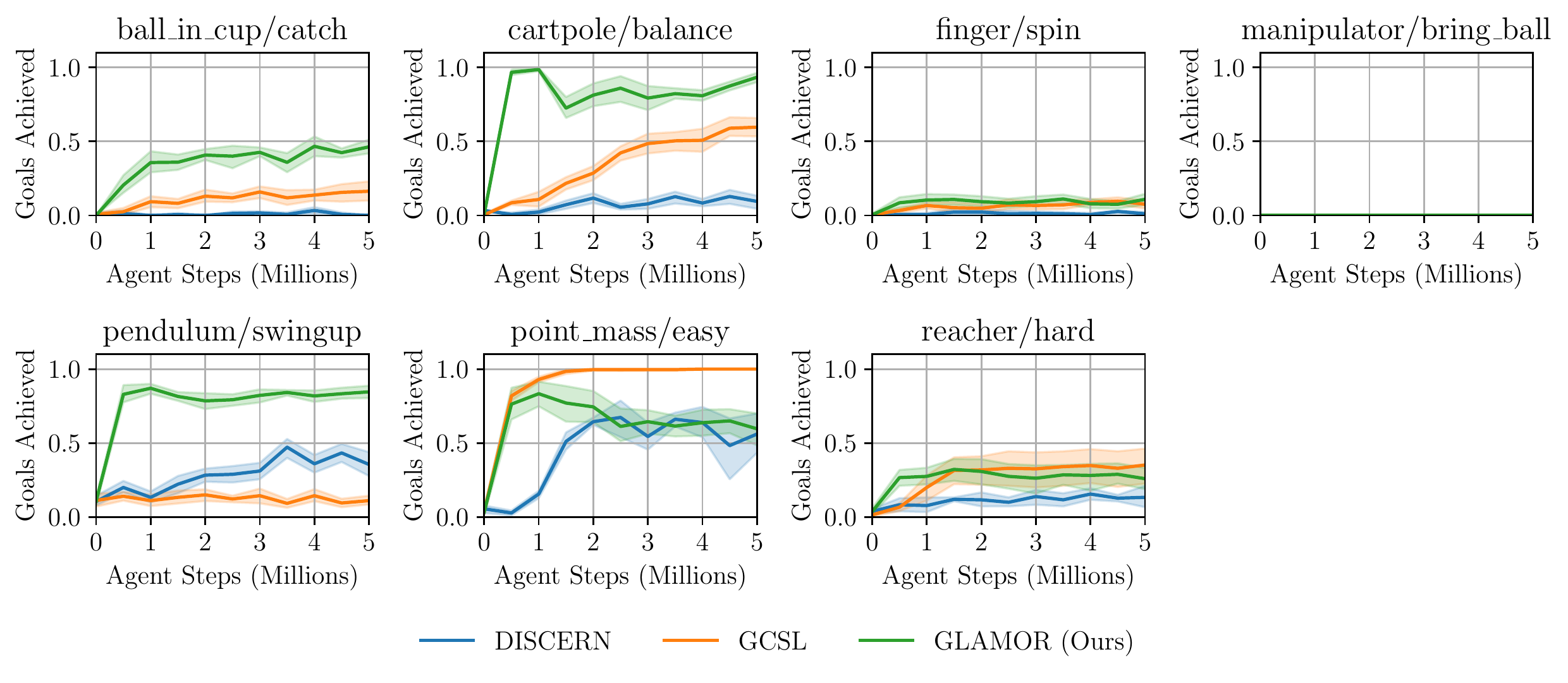}
\caption{Training Curves for Control Tasks. \algoshort{} achieves more goals (often with many fewer steps) than both GCSL and DISCERN.}
\label{fig:control_curves}
\end{figure}

\begin{figure}[t]
\centering
\includegraphics[width=0.9\linewidth]{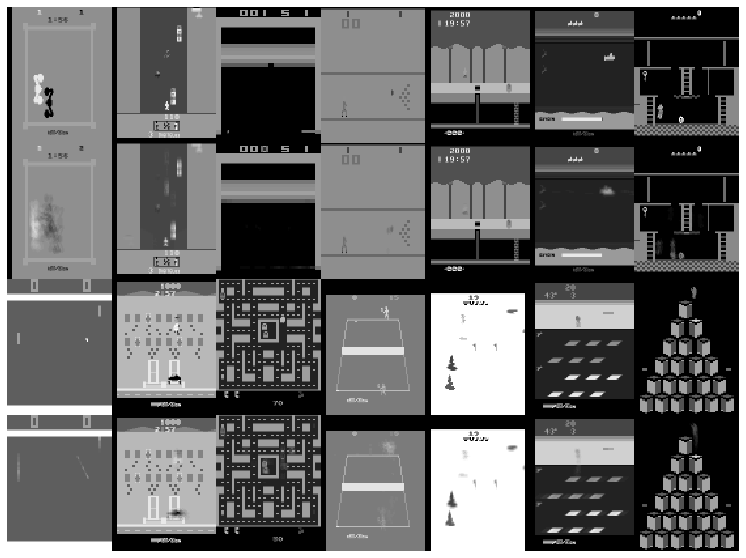}
\caption{Goal states (above) and states achieved by the fully trained \algoshort{} agent (below) averaged over 5 trials for each Atari game tested. Variance comes from environment and planning stochasticity. Note that on most games, \algoshort{} learns to control the positions of both directly and indirectly controllable objects in the frame.}
\label{fig:atari_frames}
\end{figure}

\begin{figure}[t]
\centering
\includegraphics[width=0.9\linewidth]{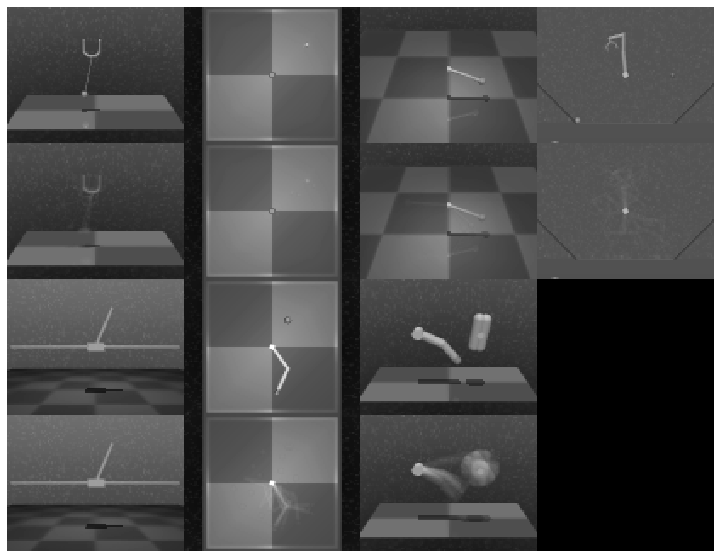}
\caption{Goal states (above) and states achieved by the fully trained agent (below) averaged over 5 trials for each control task tested. Variance comes from planning stochasticity since the environment dynamics are deterministic. In most environments, \algoshort{} learns to control the agent's state to match the visually specified goal. Note that in the finger environment, \algoshort{} learns to control the position of the finger despite not often being able to control the angle of the spinner.}
\label{fig:control_frames}
\end{figure}

In \autoref{fig:atari_curves} and \autoref{fig:control_curves}, we plot the learning curves of \algoshort{}, GCSL, and DISCERN trained for 5M agent steps. \algoshort{} learns quickly compared to the other algorithms and performs better asymptotically with the exception of a few environments (MsPacman, point-mass).
\autoref{fig:atari_frames} and \autoref{fig:control_frames} show the states achieved by \algoshort{} attemping to reach a specific goal. As noted by \citet{discern}, the \codeword{manipulator} environment is difficult and no algorithm learned to achieve goals within 5M steps. The low achievement rate on the \codeword{finger} task is due to the agent's inability to reliably control the angle of the spinner.


\end{document}